\documentclass[english]{article}
\usepackage[preprint,nonatbib]{nips_2018_wider_nonotice}

\usepackage[T1]{fontenc}
\usepackage[latin9]{inputenc}
\usepackage{color,colortbl}
\usepackage{babel}
\usepackage{verbatim}
\usepackage{url}
\usepackage{amsmath}
\usepackage{amssymb}
\usepackage{graphicx}
\usepackage{setspace}
\usepackage{hyperref}
\usepackage{cancel}
\hypersetup{unicode=true,pdfusetitle,breaklinks=false}
\usepackage[hyperpageref]{backref}
\usepackage{appendix}
\usepackage{xspace}
\usepackage{epigraph}
\usepackage[thinc]{esdiff}
\graphicspath{{figs/}}

\usepackage[labelfont=bf]{caption}
\makeatletter
\newcommand*{\rom}[1]{\expandafter\@slowromancap\romannumeral #1@}
\makeatother

\makeatletter
\newcommand{\be}{\begin{equation}}
\newcommand{\ee}{\end{equation}}

\allowdisplaybreaks \numberwithin{equation}{section}
\makeatother

\DeclareMathOperator*{\argmin}{arg\,min}
\def\eg{{\it e.g.}\xspace}
\def\ie{{\it i.e.}\xspace}

\def\crit{{\rm crit}}
\def\est{{\rm est}}

\hypersetup{
    colorlinks=true,
    breaklinks=true,
    urlcolor=blue,
    linkcolor=blue,
    bookmarksopen=false,
    filecolor=black,
    citecolor=blue,
}

\title{Unraveling the Mystery of Scaling Laws: Part \rom{1}}

\author{
    Hui Su\thanks{Equal contribution. Zhi Tian's work was done in Meituan. }
   \And
    Zhi Tian$^{\ast}$
    \And
    Xiaoyu Shen
    \And
    Xunliang Cai \AND
{Meituan Inc.}  \\
\texttt{suhui07@meituan.com} \\
}

\begin{document}
\maketitle

\begin{abstract}
Scaling law principles indicate a power-law correlation between loss and variables such as model size, dataset size, and computational resources utilized during training. These principles play a vital role in optimizing various aspects of model pre-training, ultimately contributing to the success of large language models such as GPT-4, Llama and Gemini. However, the original scaling law paper by OpenAI did not disclose the complete details necessary to derive the precise scaling law formulas, and their conclusions are only based on models containing up to 1.5 billion parameters. Though some subsequent works attempt to unveil these details and scale to larger models, they often neglect the training dependency of important factors such as the learning rate, context length and batch size, leading to their failure to establish a reliable formula for predicting the test loss trajectory.
In this technical report, we confirm that the \emph{scaling law formulations proposed in the original OpenAI paper remain valid when scaling the model size up to 33 billion}, but the constant coefficients in these formulas vary significantly with the experiment setup~\footnote{Experiments on models with larger parameters are ongoing, and we will provide the results once they are done.}. We meticulously identify influential factors and provide transparent, step-by-step instructions to estimate all constant terms in scaling-law formulas by training on models with only 1M$\sim$60M parameters. Using these estimated formulas, we showcase the capability to accurately predict various attributes for models with up to 33B
parameters before their training, including (1) the minimum possible test loss; (2) the minimum required
training steps and processed tokens to achieve a specific loss; (3) the critical batch size with an optimal time/computation trade-off at any loss
value; and (4) the complete test loss trajectory with arbitrary batch size. We further illustrate how scaling laws can aid in determining the most suitable batch/model size, dataset mix ratio and training duration under fixed computational constraints in a principled way.
Our research represents a significant shift from theoretical comprehension of scaling laws to their practical derivation and application, with the aim of advancing the development of large-scale language models. 

\end{abstract}

\newpage

\section{Introduction}
\begin{figure} 
\noindent \centering{} 
\includegraphics[width=0.48\textwidth]{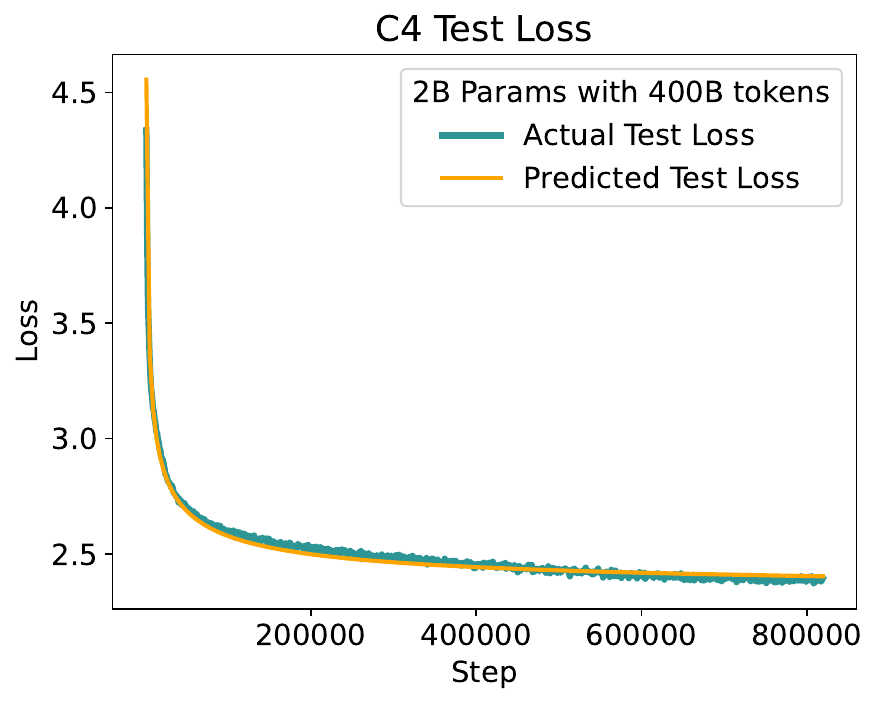} 
\includegraphics[width=0.48\textwidth]{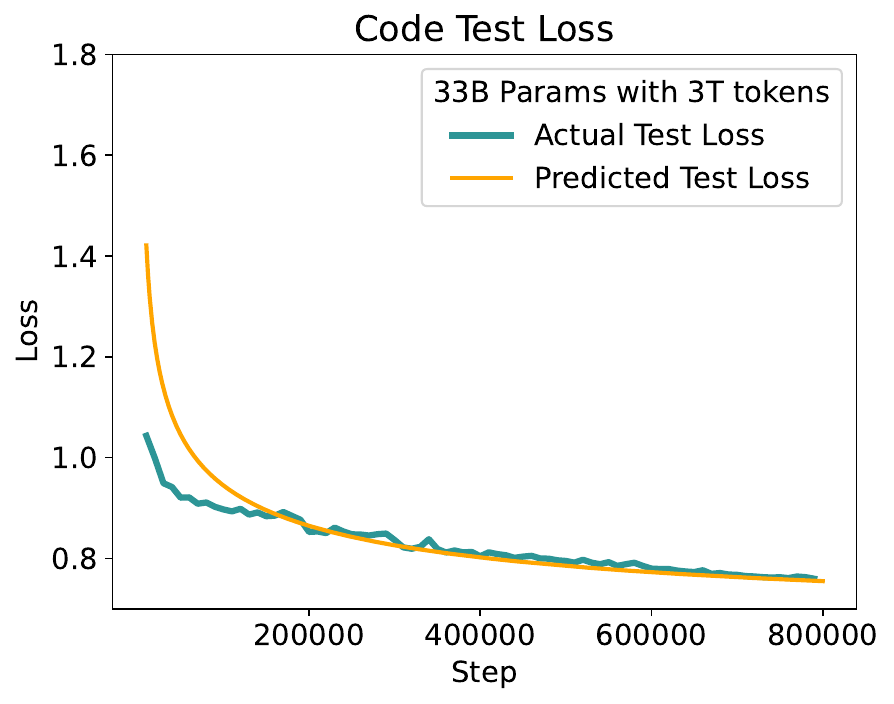}
 \caption[In Domain Test Loss]{Left: Actual and predicted loss trajectories of a 2B model on the C4 test data (Section~\ref{sec:c4_scale}). Right:  Actual and predicted loss trajectories of a 33B model on the code test data (Section~\ref{sec:mix_scale}). The actual and predicted loss trajectories closely align, especially after the initial warm-up stage.}
 \label{fig:in_domain}
\end{figure}

A wide range of studies have shown that the performance of a language model exhibits a notable growth pattern as the number of parameters and data size increase, following a power-law relationship~\cite{hestness2017deep,kaplan2020scaling,henighan2020scaling,clark2022unified,zhai2022scaling,gao2023scaling,biderman2023pythia}. 
This scaling law plays a fundamental role in the development of large language models, enabling us to estimate optimal configurations of large models from the training logs of much smaller models~\cite{tay2022scale,hoffmann2022training}. As mentioned in the GPT-4 technical report~\cite{achiam2023gpt}, some aspects of GPT-4's
performance can be accurately predicted based on models trained with no more than 1/1,000th the compute of GPT-4. By properly utilizing the scaling law, we avoid the need to perform extensive model-specific tuning on large models. 

The original scaling law paper by OpenAI presented the formulas of scaling laws and illustrated how they could aid in determining optimal training configurations~\cite{kaplan2020scaling}. Nonetheless, the presented formulas are based on static exponents estimated from their specific experiment setup. The full details on how to derive the constant terms in the scaling-law formulas for a new experiment setup (model architecture, tokenization, data distribution, etc) remain undisclosed. Furthermore, \cite{kaplan2020scaling} only conducted experiments with models containing up to 1.5B parameters, a size significantly smaller than that of contemporary large language models. There have been subsequent works that study scaling laws on larger models~\cite{clark2022unified,isik2024scaling}. Some have drawn different conclusions from the original scaling-law paper, casting doubt to the general applicability of scaling laws. For example, \cite{hoffmann2022training} claimed that the training data size should be scaled much more than the recommendation in \cite{kaplan2020scaling}. \cite{bi2024deepseek} suggested that the optimal batch size depends only on the compute budget rather than the loss value.

In this paper, we revisit the scaling-law formulas proposed by \cite{kaplan2020scaling}, confirming that they \emph{remain generally applicable when scaling the model size up to 33B}. Other works obtain different conclusions primarily due to (1) Many factors such as the data distribution, context length, tokenization affect the constant coefficients in scaling-law formulas, so the constant coefficients, unlike the formulas themselves, are not universal; and (2) The loss value adheres to an analytical power law relationship with the training step under infinite batch size. With a finite batch size, fitting the loss value with an analytical function is problematic. As a result, none of other works have provided compelling evidence to reliably predict the full loss trajectory of larger models by training solely on smaller models.

After meticulously identifying
influential factors in predicting the loss trajectory, we provide transparent, step-by-step guidelines on how to estimate all constant terms in scaling-law formulas by training on models with only 1M$\sim$60M parameters. Using these estimated formulas from small models, we showcase the capability to accurately predict various
attributes for models with up to 33B parameters before their training starts. By unravelling the mystery of scaling laws and making them easily accessible to everyone, our objective is to shift the understanding of scaling laws from theoretical concepts to practical implementation, thereby aiding future research in pre-training large language models in a more principled manner. The summary of the key results in this paper is as follows:

\begin{itemize}
    \item Hyperparameters such as batch size, learning rate, and learning rate scheduler influence the rate of convergence, yet do not impact the final converged loss provided that (1) their values fall within a reasonable range and (2) the model is trained with sufficient steps on adequate amounts of data.
    \item Adjusting the batch size involves a trade-off between time and computation. The critical batch size that strikes an optimal time/computation balance can be determined based solely on the loss value. Training with this critical batch size requires twice as many training steps to achieve a specific loss value compared to using an infinite batch size (minimum possible required steps).
    \item The context length, tokenization, data distribution and model configurations have big impacts on the constants in scaling law formulas, but do not affect the form of scaling law itself.
    \item When given a fixed context length, tokenization, data distribution, model configurations and learning rate scheduler, we observe precise and predictable power-law scalings for performance in relation to training step, batch size, and model size, provided that the learning rate is optimally configured.
    \item By training models with fewer than 60 million parameters, we can accurately estimate the constants in scaling-law formulas. This allows us to predict various attributes for models with up to 33 billion parameters before their training, including (1) the minimum possible loss; (2) the minimum required training steps and processed tokens to achieve a specific loss; (3) the critical batch size at any loss value; and (4) the complete test loss trajectory with arbitrary batch size.
    \item These predicted attributes have many intriguing features, assisting us in identifying crucial factors before training large models, such as the optimal model size and training steps within a fixed computational budget, the necessary amount of data, the ideal mix ratio of multiple datasets, and more.
\end{itemize}

\section{Preliminary}
Essentially, scaling laws~\cite{kaplan2020scaling} reveal how to predict the \textit{validation/test} loss\footnote{for simplicity, we use ``test loss'' thereafter since we observed a consistent trend for loss measured in various data distributions, differing only in the constant terms within the scaling laws.} of a given model, which can be the final loss when the model is trained to converge, or the loss at a certain step amid training. Scaling laws have been proven to be widely valid across a variety of likelihood-based tasks~\cite{henighan2020scaling}. By adhering to scaling laws, researchers can uncover patterns in how changes in model parameters and training data impact the overall effectiveness of large language models before actually training them. 

\paragraph{Notation}To enhance the clarity of explanations, we use the following notations throughout the paper, most of which are adapted from \cite{kaplan2020scaling}:
\begin{itemize}
\item $L$ -- the cross entropy loss in nats averaged over the tokens in a context
\item $N$ -- the number of model parameters, \emph{excluding all vocabulary and positional embeddings}  
 
\item $B$ -- the batch size
\item $B_{\rm crit}$ -- the critical batch size defined in \cite{mccandlish2018empirical}. Training at the critical batch size provides a roughly optimal compromise between time and compute efficiency
\item $E$ -- amount of processed tokens
\item $E_{\rm min}$ -- an estimate of the minimum amount of processed tokens needed to reach a given value of the loss. This is also the number of training steps that would be used if the model were trained at a batch size much smaller than the critical batch size
\item $S$ -- number of training steps
\item $S_{\rm min}$ -- an estimate of the minimal number of training steps needed to reach a given value of the loss.  This is also the number of training steps that would be used if the model were trained at a batch size much greater than the critical batch size
\item $N_c, S_c, B_*,\alpha_c,\alpha_S,\alpha_B$ -- constant terms in scaling-law formulas that need to be estimated
\end{itemize}

\paragraph{Goal} The original scaling-law paper is comprehensive, encompassing a wide range of content. To emphasize the key aspects useful  for pre-training large language models, this paper concentrates on estimating the following three functions, which serve as foundations of scaling laws. Using these three functions, we are able to accurately predict the training behavior of large language models before the training starts:
\begin{itemize}
\item $L(N)$ -- predict the converged loss in nats when training a model with $N$ parameters
\item $L(N,S_{min})$ -- predict the value of loss at a certain training step $S_{min}$ for a model with $N$ parameters, given that the batch size is infinite such that the number of training steps is minimized
\item $L(N, S, B)$ -- predict the value of loss at a certain training step $S$ for a model with $N$ parameters under a finite batch size $B$
\end{itemize}

\paragraph{Assumption}
To simplify the discussions, we stick to the following assumptions to derive the precise scaling laws, each of which mirrors real-world scenarios in pre-training modern large language models:
\begin{enumerate} 
    \item The model follows the standard Transformer architecture~\cite{vaswani2017attention} without obvious bottlenecks~\footnote{\eg, a limited amount of attention heads or inadequate layer normalization, which we find to make destabilize the estimation process of scaling laws.}, which has been the de facto architecture in large language models such as GPT~\cite{brown2020language}, Llama~\cite{touvron2023llama} and Gemini~\cite{team2023gemini}
    \item We assume access to an extensive set of training data and the training process never re-uses the data. This reflects the typical scenario in model pre-training since the proliferation of online platforms and digital content has contributed to the abundance of available data
    \item The training data is uniformly distributed across the training steps. This is also a reasonable assumption since the training data is often randomly shuffled in the pre-training stage, unless special tricks such as curriculum learning~\cite{bengio2009curriculum} is applied. 
   \end{enumerate}
\section{Deriving Scaling Laws}
\label{sec:derivation}
\begin{center}
\textit{``Scaling laws are decided by god; \\
The constants are determined by members of the technical staff''} 
\\ \hfill --- Sam Altman
\end{center}
In this section, we provide clear instructions on how to derive the scaling laws in a step-by-step manner. Most importantly, we unveil the full details for practically estimating all constants in scaling-law formulas, a foundational aspect emphasized by Sam Altman in the context of scaling laws.
\subsection{Predicting $L(N)$}\label{sec:test_loss_convergence}
First, let us discuss the final test loss prediction. Assume that we have infinite training data (\ie, infinite number of tokens), and we are going to train a model having $N$ parameters, scaling laws~\cite{kaplan2020scaling} draw the following correlation:
\begin{equation}\label{L_N}
    L(N) = \left(\frac{N_c}{N}\right)^{\alpha_N},
\end{equation}
where $N_c$ and $\alpha_N$ are constant scalars that can be found by statistical fitting, and $L(N)$ is the \textit{final test loss} when the model converges. In other words, $L(N)$ is the \textit{test loss limit} that an $N$-parameter model can achieve at the best (\ie, given the infinite training data, optimal optimizer and hyperparameters, and long enough training time). Note that since we assume the model is trained with infinite data, overfitting is impossible and thus the training loss should exhibit almost the same trend as the test loss.

\paragraph{Estimating $N_c$ and $\alpha_N$.} To estimate the two constant scalars, we can train a series of $k$ models with various numbers of parameters (say, $N = 1M, 10M, \dots, 10(k-1)M$) on the \textit{infinite training data} until these models converge and obtain their final test losses. With these pairs of $(N_0, L_0), \dots, (N_{k-1}, L_{k-1})$, we can obtain $k$ equations in the form of $\alpha_N \log N_c - \alpha_N \log N_i - \log L_{i} = 0 \big|_{i=0}^{k-1}$, which are linear w.r.t. $\alpha_N$ and $\log N_c$.  $N_c$ and $\alpha_N$ can then be estimated by parametric fitting using linear regression. In practice, we find that setting $k=7$ is sufficient for an accurate estimation.

\paragraph{Tolerance with Hyperparameters}
In our experiments, we find that given sufficient training data, hyperparameters such as batch size, learning rate, and learning rate scheduler influence the rate of
convergence, yet do not impact the final converged loss as long as their values fall within a
reasonable range. This finding aligns with previous research~\cite{kaplan2020scaling,bi2024deepseek}. Therefore, in order to obtain these $k$ data points of $(N_0, L_0), \dots, (N_{k-1}, L_{k-1})$, there is no need to perform extensive hyperparameter tuning. Instead, it is enough to use a fixed set of hyperparameters for all the $k$ models, provided that the model is trained with no repeated data until convergence.

\paragraph{What does infinite training data mean?} The above steps require training models with a fixed set of hyperparameters on infinite data until convergence. In practical scenarios, however, data is always finite in practice. Nevertheless, it is feasible to have \textit{relatively} ``infinite training data'' for a given model. Say we have an extremely small model with only $N= 10$ parameters. For this micro model, due to its very limited capacity, no matter how much training data we use to train it (\eg, 1B, 2B, or 1T tokens\footnote{Note that according to our third assumption, the quality of the training data is uniform and independent of the size.}), the model's performance cannot be improved, and the training dynamic (\ie, the test loss at any given training step) is almost the same. This is to say, beyond a \textit{critical size}, this micro model is unaware of the growth of the data size. In this case, we can say that the training data is \textit{relatively} infinite for this model. In fact, for a given parameter with $N$ parameters, this critical size could be computed (see Eq. 4.4 in \cite{kaplan2020scaling}). Here, we give a simple empirical method to deduce whether a data size is (relatively) infinite for a $N$-parameter model. Note that overfitting should not happen when the training data is (relatively) infinite. As a result, we can train the $N$-parameter model and observe whether the training and test losses diverge. If the dynamics of the training and test losses are almost the same everywhere (\ie, no overfitting), we can safely infer that the training data is (relatively) infinite for the given $N$-parameter model.

\subsection{Predicting $L(N, S_{min})$}
\label{sec:pred_l_n_smin}
Predicting the entire test loss curve requires estimating the test loss at any given training step. This is very challenging because there are many other factors (\eg, learning rate schedules, optimizers, and batch sizes) that significantly affect the training process. However, as shown in ~\cite{kaplan2020scaling} and ~\cite{mccandlish2018empirical}, these factors' effects can be largely eliminated if we train the model at an \textit{infinitely large batch size}, where the stochastic gradient descent (SGD) becomes gradient descent (GD). The training step at the \textit{infinitely large batch size} is denoted by $S_{\min}$ because it is the minimum possible number of training steps required to attain a certain loss. Note that the larger the training batch size, the fewer the training steps required. As a result, scaling laws~\cite{kaplan2020scaling} state that, given the infinite training data and infinitely large training batch size, the test loss $L(N, S_{\min})$ at any given step $S_{\min}$ follows:
\begin{equation}\label{L_N_S_min}
    L(N, S_{\min}) = \left(\frac{N_c}{N}\right)^{\alpha_N} +\ \ \left(\frac{S_c}{S_{\min}}\right)^{\alpha_S},
\end{equation}
where $\left(\frac{N_c}{N}\right)^{\alpha_N}$ is Eq.~\ref{L_N}, and $S_c$ and $\alpha_S$ are constant scalars to be estimated.

\paragraph{Estimating $S_c$ and $\alpha_S$.} Since the first term in Eq.~\ref{L_N_S_min} has been solved in Sec.~\ref{sec:test_loss_convergence}, for any given model with $N$ parameters, say $N = 1M$, we have $L(1M, S) = C + \left(\frac{S_c}{S_{\min}}\right)^{\alpha_S}$, where $C = \left(\frac{N_c}{1M}\right)^{\alpha_N}$ is a constant because each element in the equation is known. Now, if we can train the $1M$-parameter model with the \textit{infinitely large batch size} and an optimally set learning rate~\footnote{The learning rate is set to maximize the rate at which the training loss decreases.}, many pairs of $(L, S_{\min})$ can be obtained. By taking the logarithm on both sizes of Eq.~\ref{L_N_S_min}, we obtain equations which are linear to $\alpha_s$ and $\log S_c$. Again, we can estimate $S_c$ and $\alpha_s$ with linear regression. 
\paragraph{What does infinite batch size mean?}
It is infeasible to train a model with an infinite batch size in practice, but we can employ a ``trial and error'' method to find a sufficiently large batch size that is equivalent to the infinite batch size, which we refer to as a \textit{relatively infinite batch size}. This is based on the fact that a further increase of the sufficiently large batch size does not further reduce the number of training steps required to achieve a certain loss value, thus not altering the training loss curve. As a result, this relatively infinite batch size can be found by increasing the batch size until the training loss curve becomes stationary. We empirically found that for model sizes at the magnitude of 10M, a batch of $~$40M tokens is sufficiently large. In this way, $S_c$ and $\alpha_S$ can be estimated by the loss values and steps during the training of the 10M model at this \textit{relatively infinite batch size}.

\subsection{Predicting $L(N, S, B)$}

Finding a \textit{relatively infinite batch size} for small models is easy and we can use this trick to estimate $S_c$ and $\alpha_S$. For large models, however, training with such a \textit{relatively infinite batch size} is neither affordable nor economical. In practice, we are more interested in predicting the loss trajectory for large models under a \emph{finite batch size} $B$. 
\begin{figure}
\noindent \centering{} 
\includegraphics[width=0.68\textwidth]{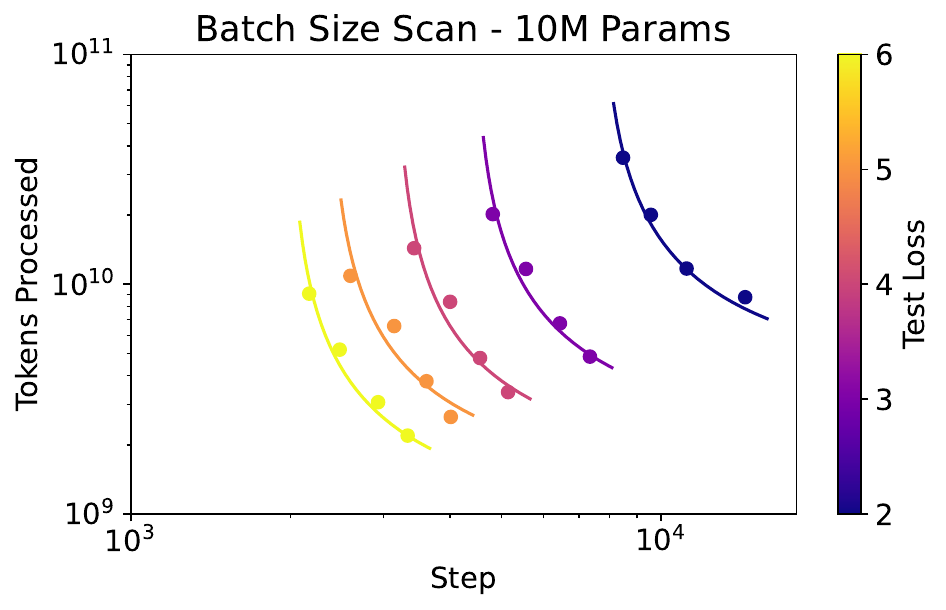} 
 \caption[Batch size scans]{Batch size scan of a 10M model with 4096 context length. Each curve has the same loss value with varying batch sizes and training steps.}
 \label{fig:batch_size_scan}
\end{figure}

\paragraph{From $S_{\min}$ to $S$.}
Thankfully, there is a conversion between the training step $S$ with any batch size $B$ and the training step $S_{\min}$ with sufficiently/infinitely large batch size. Let $\theta$ be the model parameters at some point during optimizing the model, $G_{\est}$ be the noisy gradients estimated by SGD at the point. Note that $G_{\est}$ is a random variable whose expectation is the real gradients $G$ with infinitely large batch size (\ie, $\mathbb{E}[G_{\est}] = G$). According to the Taylor expansion, the loss value after applying this parameter update is

\begin{equation}
    L(\theta-\epsilon G_{\est}) \approx L(\theta)-\epsilon G_{\est}^T G_{\est}+\frac{1}{2} \epsilon^2 G_{\est}^T H G_{\est},
\end{equation}

where $\epsilon$ is the learning rate and $H$ is the Hessian matrix. Here, the randomness introduced by $G_{\est}$ can be eliminated by computing the expectation:
\begin{equation}
\begin{split}
\mathbb{E}[L(\theta-\epsilon G_{\est})] &\approx \mathbb{E}[L(\theta)] - \epsilon \mathbb{E}[G_{\est}G_{\est}^T] + \frac{1}{2} \epsilon^2 \mathbb{E}[G_{\est}^T H G_{\est}]  \\
& = L(\theta) - \epsilon |G|^2 + \frac{1}{2} \epsilon^2 \left(G^T H G + \frac{\operatorname{tr}(H \Sigma)}{B}\right),
\end{split}
\end{equation}
where $B$ is the batch size in use, and we can obtain the decrease in the loss value is
\begin{equation}
\Delta L = - \epsilon |G|^2 + \frac{1}{2} \epsilon^2 \left(G^T H G + \frac{\operatorname{tr}(H \Sigma)}{B}\right).
\end{equation}
Note that the right-hand side is a quadratic function w.r.t. $\epsilon$, for simplicity, let $a = \frac{1}{2} \left(G^T H G + \frac{\operatorname{tr}(H \Sigma)}{B}\right)$ and $b = -|G|^2$. Therefore, the maximum decrease $\Delta L_{\max}$ is achieved when $\epsilon = -\frac{b}{2a} = \frac{|G|^2}{G^T H G + \frac{\operatorname{tr}(H \Sigma)}{B}}$, which is $\Delta L_{\max} = -\frac{b^2}{4 a} = \frac{|G|^4}{2\left(G^T H G + \frac{\operatorname{tr}(H \Sigma)}{B}\right)}$. It is worth noting that when the batch size $B \to \infty$, $\lim\limits_{B \to \infty} \Delta L_{\max} = \frac{|G|^4}{2G^T H G}$, and thus we have
\begin{equation}
    \frac{\Delta L}{\lim\limits_{B \to \infty}\Delta L} = \frac{\frac{|G|^4}{2\left(G^T H G + \frac{\operatorname{tr}(H \Sigma)}{B}\right)}}{\frac{|G|^4}{2G^T H G}} = \frac{G^T H G}{G^T H G + \frac{\operatorname{tr}(H \Sigma)}{B}} = \frac{1}{1 + \frac{\operatorname{tr}(H \Sigma)/(G^T H G)}{B}}.
\end{equation}
Let $\mathcal{B}_{\text{noise }} = \operatorname{tr}(H \Sigma)/(G^T H G)$, we have
\begin{equation}
\frac{\Delta L}{\lim\limits_{B \to \infty}\Delta L} = \frac{1}{1 + \frac{\mathcal{B}_{\text{noise }}}{B}}
\end{equation}

and thus
$\frac{\lim\limits_{B \to \infty}\Delta L}{\Delta L} = 1 + \mathcal{B}_{\text{noise }}/B$. This formulation indicates that one step with the infinitely large batch size approximately equals $1 + \mathcal{B}_{\text{noise }}/B$ steps with batch size $B$. Thus,

\begin{equation}\label{S_min_S_B_noise}
    S_{\min} = \frac{S}{1 + \mathcal{B}_{\text{noise}}/B}\
\end{equation}

\paragraph{Defining $B_{crit}(L)$}
Under the constraint of Equation~\ref{S_min_S_B_noise}, we can derive the critical batch size at $L$ which minimizes the trade-off between time ($S/{S_{min}}$) and computation (${E}/{E_{min}}$):
\begin{gather}
    B_{\rm crit}(L) = \argmin_{B} \left(\frac{S}{S_{min}} + \frac{E}{E_{min}} \right) \label{eq:crit_b_def}\\\
    E_{min} = \min_{B} {BS} = \min_{B}S_{min}(B + \mathcal{B}_{\text{noise}}) = \lim\limits_{B \to 0} S_{min}(B + \mathcal{B}_{\text{noise}}) = S_{min}\mathcal{B}_{\text{noise}} \nonumber\\
    \Rightarrow B_{\rm crit}(L) = \argmin_{B} \left( \frac{S}{S_{min}} + \frac{BS}{S_{min}\mathcal{B}_{\text{noise}}} \right) = \argmin_{B} \left(2 + \frac{\mathcal{B}_{\text{noise}}}{B} + \frac{B}{\mathcal{B}_{\text{noise}}}\right) = \mathcal{B}_{\text{noise}} = \frac{E_{min}}{S_{min}} \nonumber
\end{gather}

By substituting $B_{crit}(L)$ into Equation~\ref{S_min_S_B_noise}, we can exactly recover the formula defined in
\cite{mccandlish2018empirical}, which was shown to apply for a wide variety of neural network tasks:
\begin{gather}
S_{\text{min}} = \frac{S}{1 + B_{\crit}(L)/B}\ \label{s_min_b_crit}\\
 \Rightarrow
    \frac{BS}{B_{\text{crit}}(L) S_{\text{min}}} - 1 - \frac{B}{B_{\text{crit}}(L)} = 0 \nonumber \\
    \Rightarrow
    \frac{BS^2}{B_{\text{crit}}(L) S_{\text{min}}^2} - \frac{S}{S_{\text{min}}} - \frac{BS}{B_{\text{crit}}(L) S_{\text{min}}} = 0 \nonumber\\
   \Rightarrow \left( \frac{S}{S_{\text{min}}} - 1 \right) \left( \frac{BS}{B_{\text{crit}}(L) S_{\text{min}}} - 1 \right) = 1 \nonumber\\
\Rightarrow\left( \frac{S}{S_{\rm min}} -1 \right) \left( \frac{E}{E_{\rm min}} - 1 \right) = 1
\label{eq:TimeComputeTradeoff}
\end{gather}
We also verified the validity of Equation~\ref{eq:TimeComputeTradeoff} in our experiments (see an example in Figure~\ref{fig:batch_size_scan}). 
Putting this $B_{\rm crit}(L)$ back to Equation~\ref{eq:TimeComputeTradeoff}, we can have: 
When $B=B_{\crit}(L)$, we have $S=2S_{min}$ and $E=2E_{min}$, meaning that training with the critical batch size will cost twice the minimum number of steps and tokens necessary to reach a certain loss value $L$.

Eq.~\ref{s_min_b_crit} successfully established the conversion between $S_{min}$ and $S$ under the finite batch size $B$. In order to go from $L(N, S_{min})$ to $L(N, S, B)$, the core left is to estimate the critical batch size $B_{\crit}(L)$.

\cite{kaplan2020scaling} found that the critical batch size also roughly obeys a power law in $L$:

\begin{equation}
B_\crit(L) = \frac{B_*}{L^{1/\alpha_B}}
\label{eq:b_crit}
\end{equation}
As seen, $B_\crit(L)$ is a variable only depending on the loss value $L$. We simply need to estimate the constant terms $B_*$ and $\alpha_B$. 
\paragraph{Estimating $B_*$ and $\alpha_B$} 
In order to estimate $B_*$ and $\alpha_B$, we train a fixed-sized model with various batch sizes to generate $k$ contour lines representing consistent loss values. We then use Equation~\ref{eq:TimeComputeTradeoff} to fit these series of contour lines, yielding a set of $k$ estimated pairs $(E_{min}^1,S_{min}^1), \ldots, (E_{min}^k,S_{min}^k)$ (as shown in Figure~\ref{fig:batch_size_scan} where $k=5$ and $N=10M$). Note that Eq~\ref{eq:TimeComputeTradeoff} is linear w.r.t. $S_{min}$ and $E_{min}$, so we could use linear regression to fit them and obtain the $k$ pairs of $(E_{min},S_{min})$.

With these $k$ pairs of $(E_{min},S_{min})$, we can obtain $k$ pairs of $(B_{crit}(L), L)$ by setting $B_{crit}(L)={E_{min}}/{S_{min}}$, which can be used to estimate the values of $B_*$ and $\alpha_B$. By taking the logarithm in Eq~\ref{eq:b_crit}, the equation becomes linear w.r.t. $\alpha_B$ and $\log B_*$ and can be solved with linear regression. Empirically, we find that the value of $\alpha_B$ is quite a sensitive coefficient. Initializing it within the range of 0 and 0.5 typically results in a more stable estimation process~\footnote{As we already know the analytical function to obtain $L(N, S_{min})$ in Section~\ref{sec:pred_l_n_smin}, substituting $S_{min}$ into Eq.~\ref{s_min_b_crit} can also generate a series of numbers for $B_{crit}(L)$, which we can use to estimate $B_*$ and $\alpha_B$. Empirically we can use this method to help post-correct the estimated values, which we find to improve the estimation accuracy.}.

\paragraph{Substituting $S_{min}$ with $S$}
Finally, after establishing the analytical relationship between $S_{min}$ and $S$, we can substitute $S_{min}$ with $S$ to link $S$ with $L$.
By substituting Eq.~\ref{s_min_b_crit} into Eq.~\ref{L_N_S_min}, it yields
\begin{equation}\label{L_N_S}
\begin{split}
    L(N, S, B) &= \left(\frac{N_c}{N}\right)^{\alpha_N} +\ \ \left[\frac{S_c\cdot \left(1 + B_{\crit}(L)/B \right)}{S}\right]^{\alpha_S} \\
    &= \left(\frac{N_c}{N}\right)^{\alpha_N} +\ \ \left(\frac{S_c}{S}\right)^{\alpha_S} \cdot \left(1 + \frac{B_{\crit}(L)}{B} \right)^{\alpha_S} \\
    &= \left(\frac{N_c}{N}\right)^{\alpha_N} +\ \ \left(\frac{S_c}{S}\right)^{\alpha_S} \cdot \left(1 + \frac{B_*}{B \cdot L(N, S, B)^{1/\alpha_B}} \right)^{\alpha_S}.
\end{split}
\end{equation}
Now, let us analyze Eq.~\ref{L_N_S}, which is the relation between the loss value $L$ and the training step $S$ under a fixed finite batch size $B$. All other quantities are constant scalars that have been worked out before ($N_c,\alpha_N,S_c,\alpha_S,B_*$ and $\alpha_B$). The difficulty of computing the loss value $L(N, S, B)$ is that it appears on both sides of the equation. Although there is no analytical solution to isolate $L(N, S, B)$ here, we can numerically estimate $L(N, S, B)$ by any root-finding method (\eg, the bisection method) because it is the only unknown quantity in Eq.~\ref{L_N_S}.

Concretely, it is easy to show that Eq.~\ref{L_N_S} is monotonically decreasing w.r.t. $L(N, S, B)$:

\begin{equation*}
\begin{split}
    f (L(N,S,B)) &= \left(\frac{N_c}{N}\right)^{\alpha_N} +\ \ \left(\frac{S_c}{S}\right)^{\alpha_S} \cdot \left(1 + \frac{B_*}{B \cdot L(N, S, B)^{1/\alpha_B}} \right)^{\alpha_S} - L(N, S, B)\\
   \Rightarrow \diff{f (L(N,S,B))}{L(N,S,B)} &= -\left(\frac{S_c}{S}\right)^{\alpha_S} \left(1 + \frac{B_*}{B \cdot L(N, S, B)^{1/\alpha_B}} \right)^{\alpha_S-1}\frac{\alpha_{S}B_*}{\alpha_{B}BL(N, S, B)^{(1/\alpha_B+1})}-1<0
\end{split}
\end{equation*}

Therefore, once we find a range $(L_{\text{left}}, L_{\text{right}})$ where $f(L_{\text{left}}) \cdot f(L_{\text{right}})<0$, we can iteratively search for the critical point $L^*$ with $f(L^*)=0$ through the bisection method. In practice, the loss value is positive and falls within a restricted range, setting $L_{\text{left}}=0, L_{\text{right}}=10$ is always sufficient to cover the entire range.

\paragraph{Dependency on Learning Rate}
Eq.~\ref{L_N_S} depends only on $N, S$ and $B$ but not other hyperparameters such as the learning rate. When training with a finite batch size, the learning rate will undoubtedly have a significant impact on the loss trajectory. Removing the influence of the learning rate from Eq.~\ref{L_N_S} is clearly impossible. Empirically, we observe that the prediction from Eq.~\ref{L_N_S} is accurate when the learning rate is \emph{optimally set}, i.e., is adjusted to maximize the rate of decrease in curvature during the initial steps. As suggested in \cite{kaplan2020scaling,bi2024deepseek}, the optimal learning rate should decrease when the model size increases. We could simply adopt a ``trial and error'' method to search for the optimal learning rate in the initial training steps. Once the learning rate is determined for a fixed-sized model, we can predict its precise test loss trajectory from Eq.~\ref{L_N_S} (after the initial warm-up stage.).
\section{Experiments}
\begin{table}[h!]
\centering
\vspace{-0.5em}
\begin{tabular}{|c| c | c | c | c | c | c|} 
 \hline
Parameter & $\alpha_N$ & $\alpha_S$ & $\alpha_B$ & $N_c$ & $S_c$ & $B_*$  \\ [0.5ex] 
 \hline\hline
C4 Value  & $0.076$ & $0.67$ & $0.205$& $1.5 \times 10^{14}$ & $2.6 \times 10^{3}$  & $1.7 \times 10^{8}$ \\ \hline
\textbf{\cite{kaplan2020scaling}} & $0.076$ & $0.76$ & $0.21$ & $6.5 \times 10^{13}$ & $2.1 \times 10^{3}$  & $2.1 \times 10^{8}$ \\
 \hline
\end{tabular}
\vspace{0.5em}
\caption{The estimated constant values in scaling-law formulas on the C4 training data. The same values estimated by \cite{kaplan2020scaling} on the WebText training data are provided for comparison.}
\vspace{-1em}
\label{table:c4_para}
\end{table}

After conducting theoretical analysis and deriving scaling laws, this section presents empirical experiments to validate the efficacy of scaling laws. Following standard practice, we utilized the decoder-only Transformer architecture~\cite{vaswani2017attention} and conducted experiments on two datasets: one utilizing the C4 dataset and the other utilizing a customized mixed dataset. We followed the estimation steps outlined above to derive the scaling-law formulas.

\subsection{Scaling with C4 Dataset}
\label{sec:c4_scale}
The C4 dataset is a large, cleaned version of Common Crawl's web crawl corpus~\cite{2019t5}. The context window was set to 1024, and each batch contained about 500k tokens. We utilized 
1\% of the original C4 dataset as the test set and performed a deduplication process to ensure the removal of any overlapping text from the training set. In total, we trained only 10 small models, each with a maximum of 60M parameters, to estimate the constant terms of the scaling-law formulas. The estimated constant terms are presented in Table~\ref{table:c4_para}. The actual and predicted loss trajectories of a 2B model (30 times larger than the small model used to estimate the constant terms) using the estimated formulas are depicted on the left of Figure~\ref{fig:in_domain}. It can be observed that the predicted loss trajectory closely aligns with the actual loss trajectory. Our estimated constant terms are also not far from those estimated in \cite{kaplan2020scaling} despite using different setups, possibly due to the similar distributions of C4 and WebText, both of which consist of crawled website text. This reinforces the assertion in \cite{sharma2022scaling} that the power term in scaling laws primarily relies on the data manifold.

\begin{figure}
\noindent \centering{} 
\includegraphics[width=0.48\textwidth]{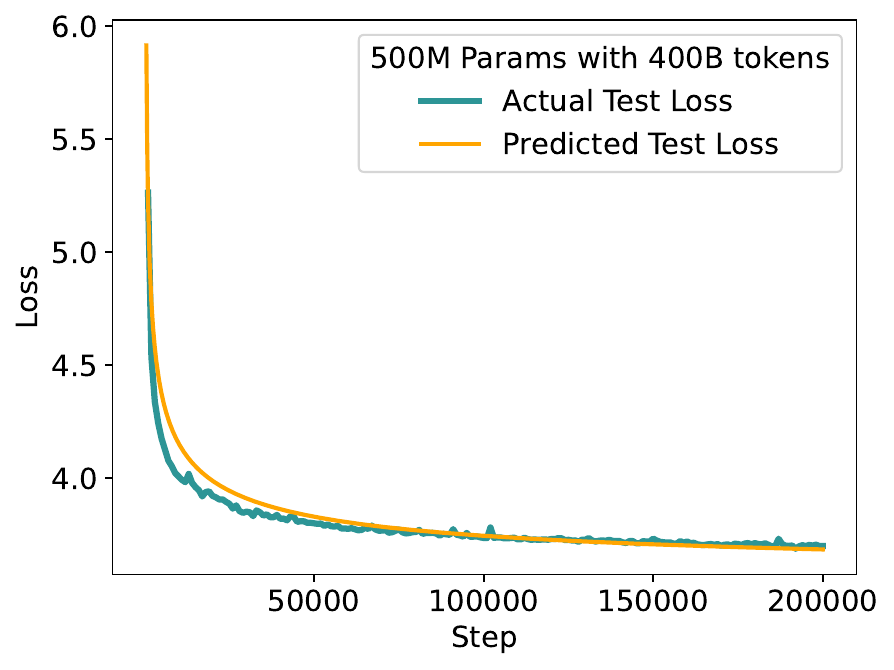} 
\includegraphics[width=0.48\textwidth]{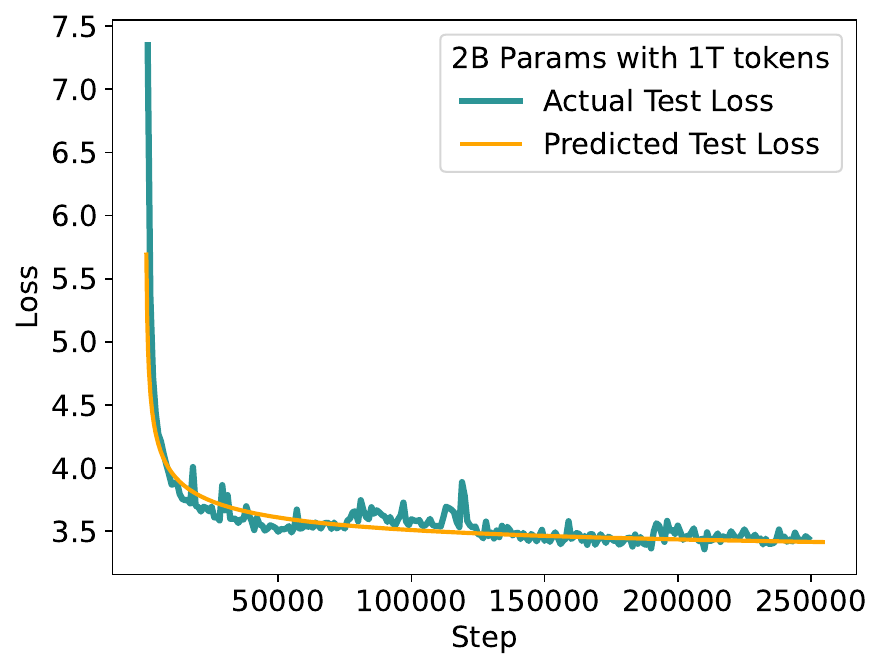}

\begin{minipage}[t]{0.48\textwidth}
    \centering
    \vspace{0pt}
    \hspace*{0.5em}
    \includegraphics[width=0.99\textwidth]{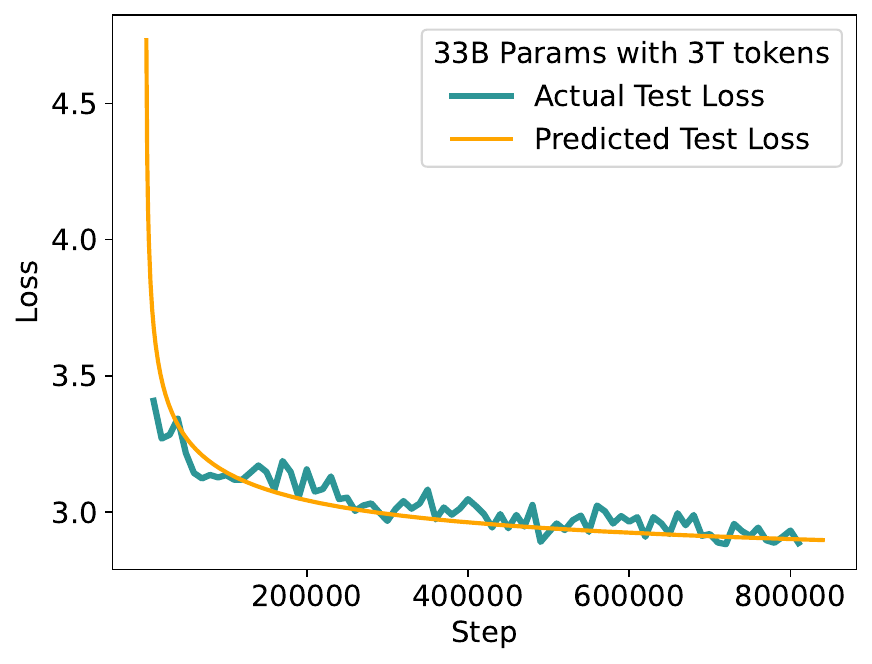}
\end{minipage}
\hfill
\begin{minipage}[t]{0.48\textwidth}
    \centering
    \vspace{4pt}
    \begin{tabular}{|c|l|}
        \hline
        Parameter  & Value\tabularnewline
\hline 
\hline 
        $\alpha_N$& $0.0615$  \\
        \hline
        $\alpha_S$ & $0.672$\\
        \hline
        $\alpha_B$ & $0.139$ \\
        \hline
         $N_c$ & $4.85 \times 10^{17}$ \\
        \hline
        $S_c$& $1.54 \times 10^{3}$  \\
        \hline
        $B_*$ & $2.15 \times 10^{11}$ \\
        \hline
    \end{tabular}
\end{minipage}

 \caption[OOD Test Loss]{Actual and predicted loss trajectories of 500M, 2B and 33B models on the out-of-domain private Chinese test data (Section~\ref{sec:mix_scale}). The loss trajectory on out-of-domain test data has large fluctuations, but the overall trend and final converged loss values still closely align with the predictions. The estimated constant values in scaling-law formulas are provided on the bottom right.}
 \label{fig:out-of-domain}
\end{figure}

\subsection{Scaling with a Large Mixed-Language Dataset}
\label{sec:mix_scale}
For the experiments with the customized mixed dataset, we manually curated a dataset containing 3T tokens comprising a mixture of English, Chinese, and code data. The data underwent a series of rigorous deduplication, filtering, and cleaning processes to ensure its quality. The context window was set to 4096, and each batch contained about 4M tokens. Similarly, we trained only 10 small models, each with a maximum of 60M parameters, to estimate the constant terms of the scaling-law formulas. The formulas are used to predict the test loss trajectory of models up to 33B (600 times larger). We test the accuracy of the predicted loss trajectory on both in-domain and out-of-domain test data.
\paragraph{In-Domain Test Loss Prediction}
For the in-domain test set, we use the code data following the same distribution as that used in the training data (the code data comprises $10\%$ of the full training data). The actual and predicted loss trajectories of a 33B model using the estimated formulas are depicted on the right of Figure~\ref{fig:in_domain}. We can see that the loss trajectory is generally accurate after 200k steps. After 200,000 steps, the predicted loss and the actual value are very accurate, but in the earlier stages, the prediction may not be as accurate due to the influence of warm-up and the large prediction multiplier causing errors. 
\paragraph{Out-of-Domain Test Loss Prediction}
For the out-of-domain test set, we use a private Chinese data whose type is very rare in the training data and can be considered as out-of-domain data. The estimated constant terms, together with the actual and predicted loss trajectories of 500M, 2B and 33B models using the estimated formulas are depicted in Figure~\ref{fig:out-of-domain}.  It is evident that predicting out-of-domain data is more challenging than predicting in-domain data, as the actual loss trajectory exhibits significant fluctuations. Nonetheless, the overall trend of actual and predicted loss trajectories closely aligns. The final converged loss values are also rather similar, affirming the efficacy of scaling laws in predicting the loss trajectory for both in-domain and out-of-domain data.

\section{Discussions}
The significance of scaling laws extends beyond mere prediction of the loss trajectory. More importantly, they can aid in pinpointing the optimal experimental configuration without requiring extensive tuning on very large models, thereby transforming the training of large language models from an alchemy-like trial-and-error process into a principled methodology. In this section, we highlight main benefits of scaling laws and discuss ways to further advance beyond them.
\paragraph{Determining $B$}
As long as all hyperparameters are well-tuned (especially the learning rate and regularization hyperparameters) and the number of training steps is sufficient, it is believed that the same final performance should be attainable using any batch size~\cite{shallue2019measuring}, so the batch size mainly influences the training speed of language models. Often, when training large language models, the ideal batch size is suggested to be set as the largest batch size supported by the available hardware
~\cite{tuningplaybookgithub}, so as to maximize the training speed without considering the computational cost.
In Eq~\ref{eq:b_crit}, we show that the critical batch size with the optimal speed/computation trade-off can be analytically computed from the loss value. Under the guidance of this formula, we would be able to estimate the preferred batch size under any loss trajectory. Furthermore, this optimal batch size in Eq~\ref{eq:b_crit} is determined by equally minimizing the training time and required computation, as shown in Eq~\ref{eq:crit_b_def}. In practice, if we would like to prioritize one over the other, we can follow the same process to derive the optimal batch size. By this means, we are able to obtain the optimal batch size based on our customized need in a systematic way.

\paragraph{Determining $N$ and $S$}
In practice, we often opt for the largest affordable model size and train the model until convergence. Nevertheless, this simplistic approach can deviate significantly from the optimal configuration and result in substantial resource wastage. Scaling laws provide a principled approach to choosing the optimal model size $N$ and number of training steps $S$ given a fixed computational budget $C$~\footnote{We follow \cite{kaplan2020scaling} to use $C \approx 6NBS$ here.}. Given that Eq~\ref{L_N_S} already provides the precise relation between the loss $L$, batch size $B$, model size $N$ and training steps $S$, we could find the model size that minimizes $L$ under the critical batch size ($B = B_{crit}$). This optimal $N$ can be obtained by taking the derivative of Eq~\ref{L_N_S} w.r.t. $N$ and setting it to $0$. By inserting this optimal $N$ into Eq~\ref{L_N_S} and eliminating the loss term, we have:
\begin{gather}
       N(C) ={N_{c}}\left(\frac{C}{C_{c}}\right)^{\alpha_{C}/\alpha_{N}}\left(1+\frac{\alpha_{N}}{\alpha_{S}}\right)^{1/\alpha_{N}}\nonumber\\
       S\left(C\right) =\frac{C_{c}}{6N_{c}B_{\ast}}\left(1+\frac{\alpha_{N}}{\alpha_{S}}\right)^{-1/\alpha_{N}}\left(\frac{C}{C_{c}}\right)^{\alpha_{C}/\alpha_{S}}\nonumber\\
L\left(N\left(C\right),C, S(C)\right)=\left(1+\frac{\alpha_{N}}{\alpha_{S}}\right)L\left(N(C),\infty\right) \label{eq:opt_loss}\\
       C_{c} =6N_{c}B_{\ast}S_{c}\left(1+\frac{\alpha_{N}}{\alpha_{S}}\right)^{1/\alpha_{S}+1/\alpha_{N}}\left(\frac{\alpha_{S}}{\alpha_{N}}\right)^{1/\alpha_{S}}\nonumber\\
       \alpha_{C} = 1/\left(1/\alpha_{S}+1/\alpha_{B}+1/\alpha_{N}\right)\nonumber
\end{gather}
where $N(C)$ and $S(C)$ are the optimal model size and number of training steps given a fixed computational budget $C$. $L\left(N\left(C\right),C, S(C)\right)$ is the final loss value with the chosen $N(c), C$ and $S(C)$.
The detailed derivation can be found in Appendix B.1 of \cite{kaplan2020scaling}. All the constant terms mentioned above are already known through the derivation steps described in Section~\ref{sec:derivation}, so we could directly estimate $N(C)$ and $S(C)$ from from our computational budget $C$. 
Note that, as shown in Eq~\ref{eq:opt_loss}, the final loss is $\alpha_N/\alpha_S$ more than the converged loss $L(N, \infty)$. Therefore, optimally we should \emph{NOT} train the model until convergence, which contrasts with the current common practice.

\paragraph{Determining Computational Budget}
In many downstream applications, we might not be right provided with a fixed computational budget. Instead, there is often a minimum threshold requirement that must be met before implementation. In such cases, we need to figure out the minimum possible computational budget in order to meet this threshold requirement. As the evaluation criteria is often correlated with the loss value, we can link this minimum threshold requirement into a certain loss value. From this loss value, we can readily determine the optimal model size and minimum computational budget required to achieve it from the analytical relation provided in Equation~\ref{eq:opt_loss}.

\paragraph{Determining Data Mix Ratio}
The quality of pre-training datasets is the one of the most important factors that affects the quality of large language models~\cite{su2022welm}. However, determining the optimal mix ratio from multiple data sources is an extremely challenging task as it involves combinatorial combinations~\cite{xie2024doremi}. Existing works usually determine domain weights (the sampling probabilities for each domain) by using intuition or a set of downstream tasks. Scaling laws can offer some new insights in helping determine the optimal mix ratio. By predicting the test loss trajectory of large models on each individual data source, we could implicitly infer how important and useful each data source is (\eg, if the loss decreases faster in one data source and converges into a lower loss value, then this data source might be more useful).

\paragraph{Context Length}
As mentioned, the context length significantly influences the values of the constant terms in scaling-law formulas. Anchoring all constant terms to a specific context length means that we need to rerun the estimation process for every new context length, which is rather inefficient because it is common to adjust the context length to fit various tasks. Given that the loss value at each position also approximately follows a power-law relation~\cite{kaplan2020scaling}, it would be possible to include the context length directly as a parameter of the formulas.

\paragraph{Mixture-of-Experts}
The mixture-of-experts (MoE) architecture has gained popularity with demonstrated superior performance compared to its dense counterpart~\cite{jiang2024mixtral}. It would be highly beneficial to derive a similar scaling law applicable to the MoE architecture. In MoE architectures, each input interacts with only a subset of the network's parameters -- chosen independently for each datapoint~\cite{denoyer2014deep,bengio2016conditional}. This changing of the architecture would inevitably impact the form of $L(N)$ because both the number of both activated and total parameters influence the loss values~\cite{clark2022unified}. The following steps, such as Eq~\ref{eq:crit_b_def} and  Eq~\ref{s_min_b_crit} are general and should not be affected.
\bibliographystyle{halpha}
\bibliography{main}

\end{document}